\documentclass[10pt,journal,compsoc]{IEEEtran}
\usepackage{graphicx}
\usepackage{comment}
\usepackage{amsmath,amssymb} 
\usepackage{color}
\usepackage{subfigure}
\usepackage{float}
\usepackage{url}
\usepackage{csquotes}
\usepackage{hyperref}

\begin{document}
%
\title{IllumiNet: Transferring Illumination from Planar Surfaces to Virtual Objects in Augmented Reality}
%
%
%
%

\author{Di Xu,
        Zhen Li,
        Yanning Zhang, Qi Cao
\IEEEcompsocitemizethanks{\IEEEcompsocthanksitem Di Xu is with the AI Lab of Shadow Creator Inc., Beijing,
China.\protect\\
E-mail: di.xu@ivglass.com
\IEEEcompsocthanksitem Zhen Li and Yanning Zhang are with School Computer Science, Northwestern Polytechnical University, Xi'an, China.
\IEEEcompsocthanksitem Qi Cao is with the School of Computing Science, University of Glasgow, Singapore campus.}
}

\IEEEtitleabstractindextext{%
\begin{abstract}
This paper presents an illumination estimation method for virtual objects in real environment by learning. While previous works tackled this problem by reconstructing high dynamic range (HDR) environment maps or the corresponding spherical harmonics, we do not seek to recover the lighting environment of the entire scene. Given a single RGB image, our method directly infers the relit virtual object by transferring the illumination features extracted from planar surfaces in the scene to the desired geometries. 
Compared to previous works, our approach is more robust as it works in both indoor and outdoor environments with spatially-varying illumination. Experiments and evaluation results show that our approach outperforms the state-of-the-art quantitatively and qualitatively, achieving realistic augmented experience.
\end{abstract}

}

\maketitle

\IEEEdisplaynontitleabstractindextext

%
\IEEEpeerreviewmaketitle

\IEEEraisesectionheading{\section{Introduction}\label{sec:introduction}}

%
%
%
%

\IEEEPARstart{C}{ompositing} rendered virtual objects into real scenes is a fundamental but challenging problem. Emerging applications such as augmented reality (AR), live streaming, and film production all demand realistic compositing.
The high dynamic range (HDR) environment maps are usually adopted to record the illumination of the entire scene. It reproduces a great dynamic range of luminosity  which is even higher than that of the human visual system. However direct capture of HDR images is not feasible for most cases, as it requires tedious set-ups and expensive devices~\cite{reinhard2010high}.
On the contrary, commercial augmented reality tools, e.g. Apple's ARkit, provide lightweight mobile applications to estimate the scene illumination. But these techniques only consider the camera exposure information and are actually quite rudimentary.

\begin{figure}
	\centering  
	\subfigure{\includegraphics[width=0.45\textwidth]{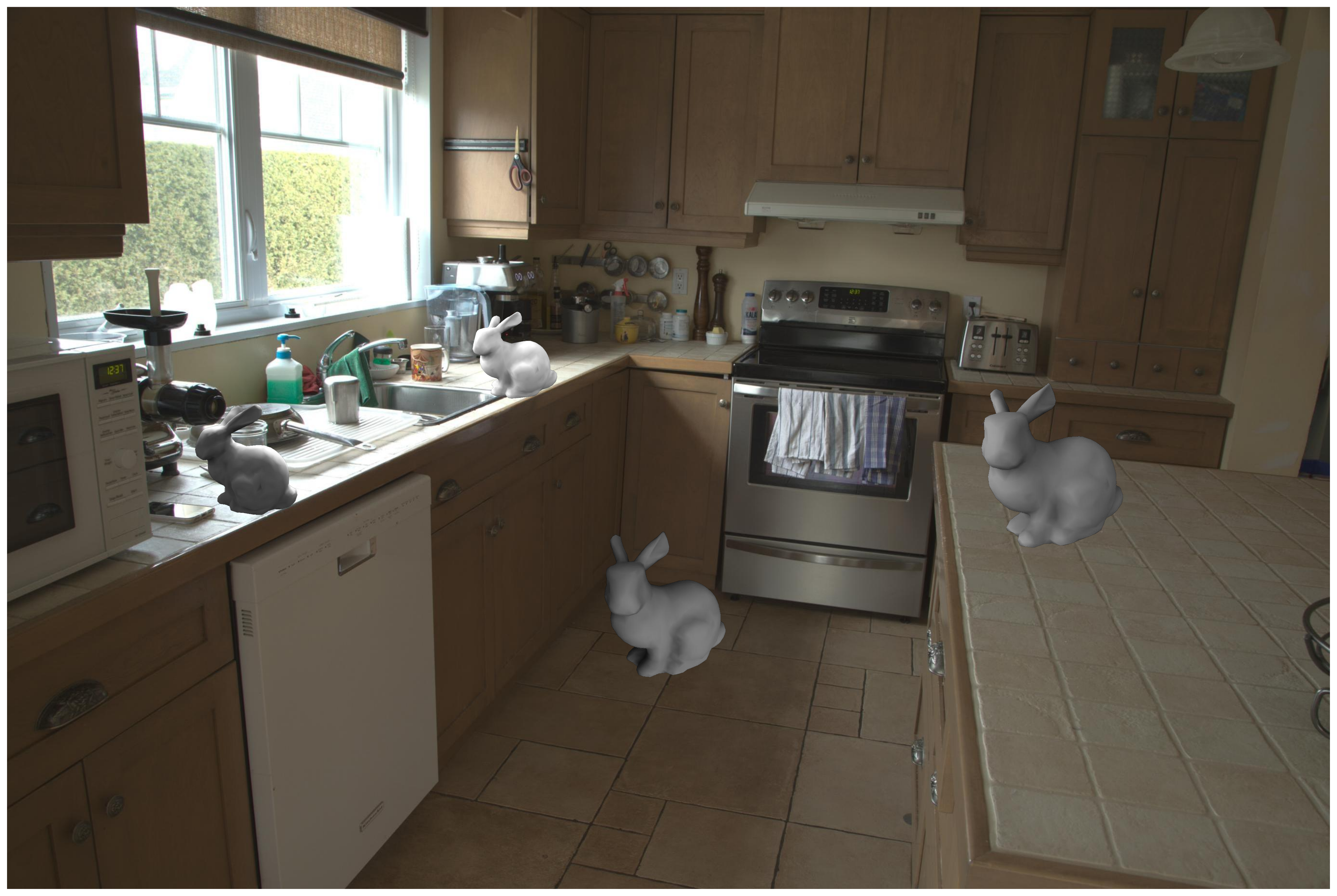}}
	\caption{Given a single RGB image, our method directly infers the rendered virtual object by transferring the illumination features of planar surfaces in real scenes, without recovering an environment map. It generates realistic rendering with spatially-varying illumination.}
	\label{fig:starter}
\end{figure}

In order to achieve realistic rendering, previous approaches try to obtain the HDR environment maps in various ways. For example, some works propose to insert certain objects to the scene, such as light probes~\cite{Debevec98,Debevec2012}, 3D objects~\cite{georgoulis2017around,weber2018learning} with known properties, or human faces~\cite{Yi_2018_ECCV,Calian18}. Some assume they have additional information e.g. panoramas~\cite{zhang2017learning}, depth~\cite{maier2017intrinsic3d}, or user input~\cite{karsch2011rendering}. Although these methods work well in certain scenarios, such requirements are usually not feasible for practical applications. Therefore, recent works try to infer the HDR environment maps from limited input information by learning. For example~\cite{gardner-sigasia-17,Song_2019_CVPR,Garon_2019_CVPR} propose to recover the HDR environment maps from a single limited filed-of-view (FOV) low dynamic range(LDR) image for indoor scenes, while~\cite{hold2017deep,Zhang_2019_CVPR,hold2019deep} take use of the sky model and try to infer the outdoor lighting. Although the learning-based methods achieve plausible results, recovering the illumination of the entire scene is still a highly ill-posed problem, mainly because of the complexity of HDR environment maps and the missing information from the input LDR image. The illumination of a scene is a result of many factors including various lighting sources, surface reflectance, scene geometry, and object inter-reflections. The limited FOV, which only captures 6\% of the panoramic scene according to~\cite{Chloe19}, makes the problem even harder since the light sources are very likely not captured in the input image. More importantly, HDR environment maps only account for the illumination incident from every direction at a particular point in the scene, which is often violated for spatially-varying lighting in the scene~\cite{gardner2019deep}. It means that, describing illumination of the entire scene with a single HDR environment map may fall short for realistic rendering of virtual objects. 

Therefore in this research project, we propose an object illumination estimation method by transferring the lighting conditions from 3D planes detected in real scenes to some kind of virtual object. Rather than learning a HDR environment map, we directly infer the relit virtual object itself. On the one hand, the per-vertex lighting model from~\cite{xu2018shading}, overall illumination(OI),are utilized, representing the overall effect of all incident lights at the particular 3D point. On the other hand, planar regions are quite common in both indoor and outdoor scenes. They offer important geometric and photometric cues in tasks such as scene reconstruction~\cite{chauve2010robust}, navigation~\cite{liu2019planercnn}, and scene understanding~\cite{tsai2011real}. Taking advantage of the easy-to-obtain OI from the planes in the scene, we propose a novel generative adversarial network (GAN) to transfer the deep feature. 
We use common objects first such as planes and some kind of virtual object to train an autoencoder network to learn deep features from OI. Then we guide the GAN to transfer OI between different objects with the photo-consistency constraint. Finally, relighting objects are generated from the predicted OI.

The main contributions of the proposed method are as follows:

\begin{enumerate}
	\item Rather than recovering the complicated HDR environment map from a single RGB image, we propose a novel framework that directly infers the rendered virtual object by transferring the illumination features of planar surfaces in real scenes.
	
	\item Our method is robust to handle both indoor and outdoor scenes with spatially-varying illumination, which is more versatile than previous approaches~\cite{gardner2019deep,gardner-sigasia-17,Garon_2019_CVPR,hold2019deep,hold2017deep,Song_2019_CVPR,Zhang_2019_CVPR} only focusing on a particular case.
	\item Although the proposed GAN framework is trained with planar surfaces in this paper, it's also feasible for other common geometries for illumination transfer. 
	
\end{enumerate}

\begin{figure*}
	
	\begin{center}
		\includegraphics[width=0.97\textwidth]{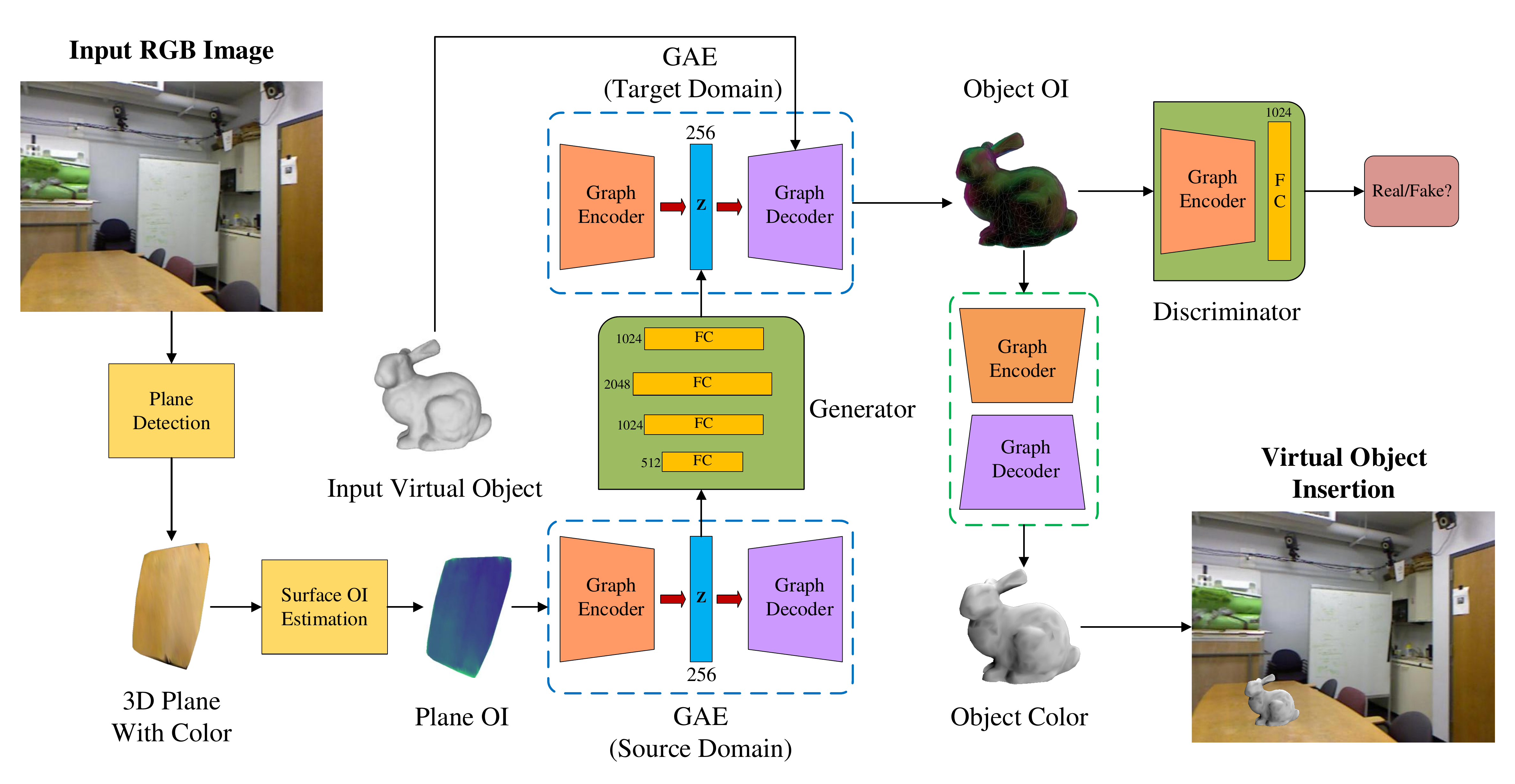}
	\end{center}
	\caption{Our framework. 
		Taking a single RGB image as the input, we detect the planar surfaces in the scene and compute its overall illumination(OI)~\cite{xu2018shading}. The illumination deep features are extracted by a graph autoencoder(GAE), and then transferred to the virtual object using the proposed GAN. Finally, the virtual object is inserted to the image according to the predicted OI.}
	\label{fig:framework}
\end{figure*}

\section{Related Works}
Illumination estimation from a scene is a long-standing topic and has been extensively studied. The problem is complicated, even ill-posed sometimes, since it depends on multiple factors, including lighting, scene geometry, surface material, reflectance, etc. Direct capture methods were first proposed by Debevec~\cite{Debevec98,Debevec2012} by taking photographs of a polished metal ball in the scene. An omnidirectional HDR radiance map with great dynamic range of luminosity was then reconstructed, and could be used to render virtual objects into the scene. However inserting such an additional tool into the scene is infeasible for most scenarios and difficult to scale. Other than HDR environment map, spherical harmonics (SH) were often used to parameterize incident illumination~\cite{barron2013intrinsic,wu2011high,johnson2011shape,wu2014real}. 
Due to the high computational cost of SH, usually only the low-frequency part was used during the optimization, e.g. 2nd-order SH in~\cite{wu2014real} and 5th-order in~\cite{Garon_2019_CVPR}. Even though the use of SH could simplify the formulation of incident illumination, it still required the computation of visibility map and the estimation of albedo. It actually took a considerable amount of time especially for a dense mesh~\cite{wu2011high}. Therefore in~\cite{xu2018shading}, rather than recovering the lighting of the whole scene, Xu et al. introduced a novel term named vertex overall illumination vector to represent the overall effect of all incident lights at each individual 3D point of the object. However its improvement over SH was only showed in term of shape-from-shading. Meanwhile, lighting estimation is also studied as an intermediate result for specific purposes. For example in \cite{gao2017naturalness} and \cite{pei2017color}, authors estimated the lighting for the purpose of image enhancement. While~\cite{han2019asymmetric} proposed to normalize the illumination on human faces, in order to improve the performance of face recognition.

Thanks to the rapid development of deep learning, recent works tried to directly estimate illumination from a single LDR image with limited FOV. Garder et al.~\cite{gardner-sigasia-17} first proposed an end-to-end CNN to recover environment maps from a single view-limited LDR image in an indoor scene. Their approach first used a large number of LDR panoramas with source light position labels to train and predict the position of the light source, and then used a small number of HDR panoramas to fine-tune the network to estimate light intensity. Hold-Geoffroy et al.~\cite{hold2017deep} learned to predict the parameters of the Hosek-Wilkie sky model from a single image to get the outdoor scene illumination. Zhang et al.~\cite{Zhang_2019_CVPR} used a more sophisticated Lalonde-Matthews(L-M) outdoor light model to predict model parameters from LDR images for an outdoor HDR panorama. Cheng et al.~\cite{Cheng2018} proposed utilizing two pictures taken from the front and back of the phone to estimate low-frequency lighting. ~\cite{Chloe19} used a special camera device to take scene photos and polished steel balls of three different materials to collect pairs of image and HDR environment map data. Calian et al.~\cite{Calian18} used the Sun+Sky model and face prior to estimate the HDR light probe from the LDR face, but it is prone to local minima. When the light source is behind the person, the model estimates the wrong result because it is unable to get enough illumination information from the backlit face. Song et al.~\cite{Song_2019_CVPR} designed three sub-neural networks to progressively estimate geometry, LDR panoramas and final HDR environment map based on input image and locale. Garon et al.~\cite{Garon_2019_CVPR} proposed a method for estimating the spatially-varying indoor illumination in real time, which combines global features and local features to predict spherical harmonics coefficients. However, due to the complexity and unknownness of the real scene, especially when the light sources are not captured in the input image, inconsistent illumination of predicted panoramic HDR is inevitable. Essentially, these previous works get the mapping of input images to environment maps or SH coefficients. While our proposed approach directly predicts the illumination effects of the inserted virtual object itself, making the problem less error-prone.

In their recent work~\cite{gardner2019deep}, Gardner et al. proposed to replace the HDR environment maps with parametric representations. The idea is somewhat similar to ours, but there are two major differences. Firstly their lighting model is a set of discrete 3D lights describing the entire science, while our proposed approach directly transfer the vertex overall illumination from detected planes to the virtual objects. Secondly, as indicated in its title, the method in~\cite{gardner2019deep} only applies to the indoor illumination, which is less robust compared to our method that works for both indoor and outdoor scenes. Some recent works also proposed to estimates the HDR lighting environment maps from more complicated inputs. For example, Gkitsas et al. presented a data-driven model that estimates lighting from a spherical panorama~\cite{gkitsas2020deep}. Srinivasan et al. proposed to estimate a 3D volumetric RGB¦Á model, and then the
incident illumination, using narrow-baseline stereo pairs of images~\cite{srinivasan2020lighthouse}. While their methods both achieved realistic results, the inputs are usually more difficult to obtain compared to standard RGB images.

\section{Proposed Method}

In our research, the goal is to transfer the lighting effects of common structures in the scene, i.e. planar surfaces, to the inserted virtual objects. As shown in Fig.~\ref{fig:framework}, 
for an input RGB image, we first detect the planar surfaces of any specific region for virtual object compositing. The OI of the particular plane is then calculated. After that a graph autoencoder (GAE)\cite{kipf2017semi-supervised} is applied to extract the corresponding deep feature, which is then transferred from planes to the virtual object. Finally, the rendered color of the virtual object is obtained from its corresponding OI. 
In the remaining parts of this section, we will illustrate in details of each sub-module, network architecture, as well as the implementation details.
%

\subsection{Overall Illumination and Plane Detection}

\begin{figure*}
	\centering  
	\subfigure[]{
		\includegraphics[width=0.35\textwidth]{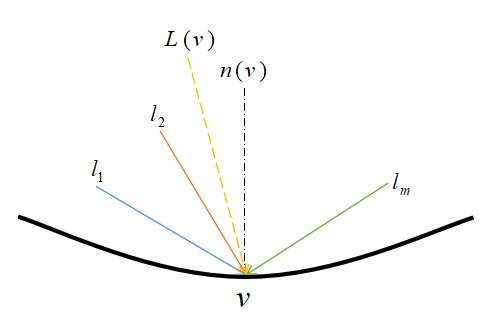}}
	\subfigure[]{
		\includegraphics[width=0.6\textwidth]{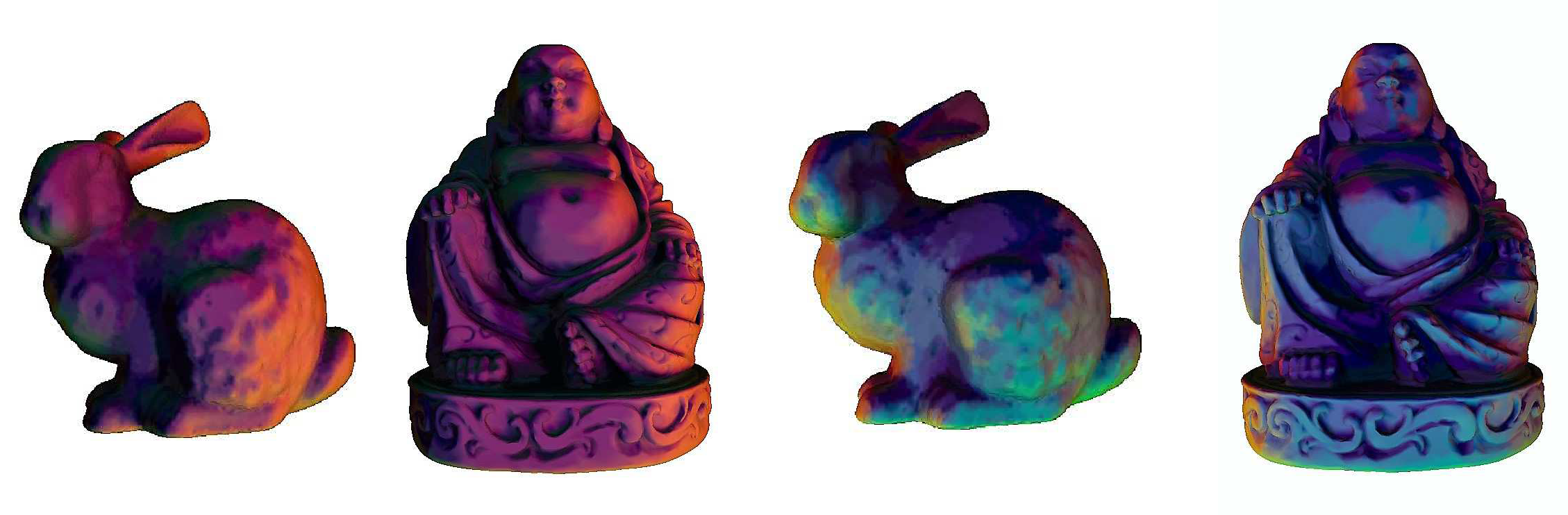}}
	\caption{(a)At each point $v_i$ on the surface, the vertex overall
		illumination vector $L(v_i)$ represents the overall effect of all
		incident lights such as $l_1, l_2, \cdots, l_m$ from different
		directions. 
		(b)Visualization of $L(v)$ on 3D models, where the 3D vector is directly mapped to RGB color space.}
	\label{fig:probe}
\end{figure*}

The concept of vertex OI was first proposed in~\cite{xu2014recovering}, describing the overall effect of all incident lights at each point of the object. As shown in Fig.~\ref{fig:probe}(a), for a vertex $v$ on a 3D model, the 3D vector $L(v)$ is denoted as its OI, and {\bf n}(v) denoting as its unit surface normal. Then the reflected radiance of $v$ can be computed as: 
$I(v)  = L(v) \cdot {\bf n}(v)$. 
In~\cite{xu2018shading} the OI was applied on Debevec's light probe images~\cite{debevec2008rendering}. A light probe image is an omnidirectional, high dynamic range image that records the incident illumination conditions at a particular point in space. Such images are usually captured under general and
natural illumination conditions. Therefore in this research project, instead of estimating individual light sources in the scene and computing the visibility function of each vertex, we infer the OI of each vertex for the purpose of object compositing. Some examples of OI of relit 3D models are shown in Fig.~\ref{fig:probe}(b), where the 3D vectors of OI are mapped to RGB color for a better understanding. More details about overall illumination can be found in the supplementary material.

\begin{figure*}
	\begin{center}
		\includegraphics[width=0.8\textwidth]{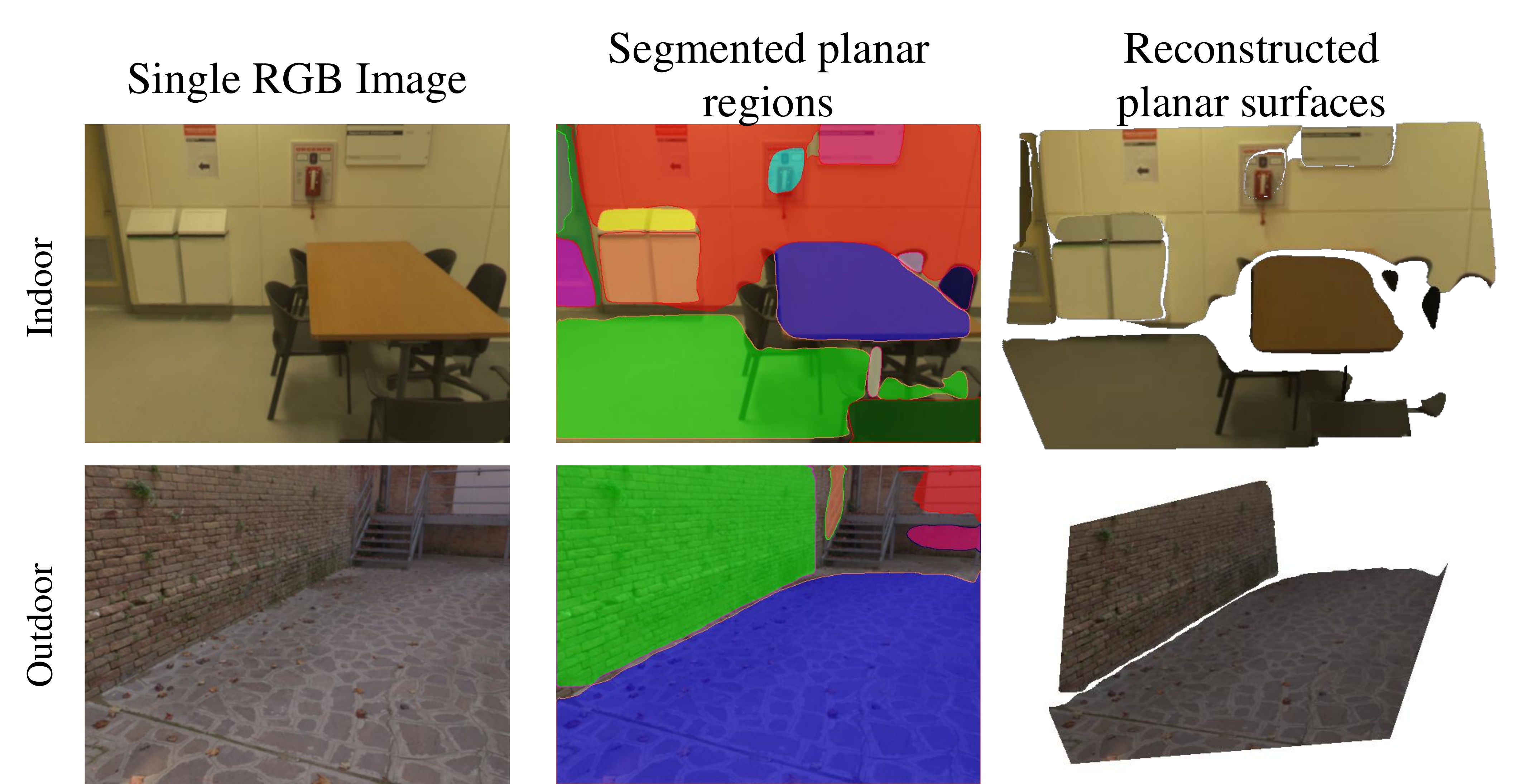}
	\end{center}
	\caption{Plane detection results using~\cite{liu2019planercnn}. }
	\label{fig:planercnn}
\end{figure*}

Different from previous methods that require to insert certain geometries~\cite{Calian18,Debevec98,Debevec2012,georgoulis2017around,weber2018learning,Yi_2018_ECCV}, we make use of the \emph{planes} that already exist in the scene. Planar surfaces with different sizes and shapes, e.g. floors, walls, tables or the ground, are some of the most commonly seen geometries in all kinds of indoor and outdoor scenes.  Therefore given a single RGB image, we first detect the planes using the existing method~\cite{liu2019planercnn}. As shown in Fig.~\ref{fig:planercnn}, it reconstructs 3D piecewise planar surfaces and estimates the corresponding 3D coordinate from a single RGB image. Inside the region for virtual object compositing, planes with appropriate size and orientation are selected. Then the OI of the plane can be calculated according to~\cite{xu2018shading}. Since the OI depicts the illumination property of the 3D model in a particular scene, we are able to transfer this property from one model to another in a learning-based manner.

\subsection{Extracting Deep Features}\label{sec:GAE}
\begin{figure*}[t]
	\begin{center}
		\includegraphics[width=0.95\textwidth]{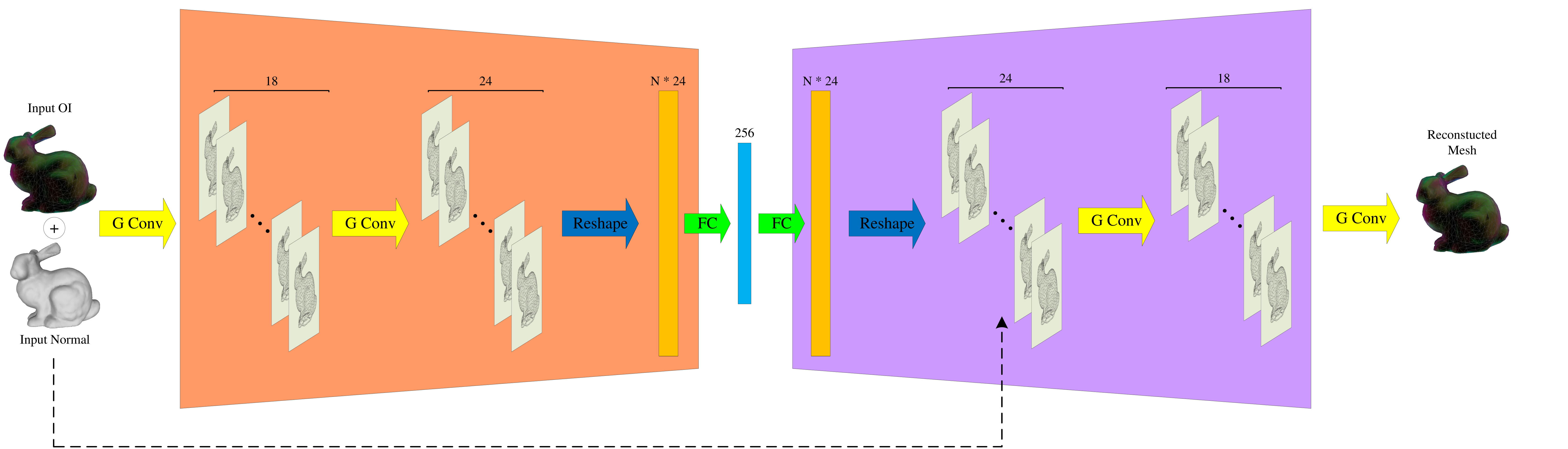}
	\end{center}
	\caption{Structure of our graph autoencoder, which is used to extract the deep feature for transferring illumination. }
	\label{fig:gae}
\end{figure*}

Before transferring the illumination, we need to extract its deep features first. Inspired by the graph convolution network proposed in~\cite{kipf2017semi-supervised}, we design a GAE structure, which is independently trained to learn the OI feature representation of the source object and the target object. As shown in Fig.~\ref{fig:gae}, the encoder-decoder consists of a two-layer graph convolution and a Fully Connected(FC) layer. The latent feature vector is 256-dimensional. Each GAE contains a unique representation of this domain object, including shapes, normals, poses, etc.. Since the input data to our network are 3D models, we define our graph as an undirected graph $G=(V,E)$, where E is the adjacency matrix of the graph, and $V$ is the feature matrix with a dimension of 6 times of the vertex number, including normal, OI / RGB information. According to~\cite{kipf2017semi-supervised}, the single layer of the graph convolutional neural network is defined in Eq.(1).:

\begin{equation}
H^{(l+1)} = \sigma{(\tilde{D}^{-\frac{1}{2}}\tilde{A}\tilde{D}^{-\frac{1}{2}}H^{(l)}W^{(l)})}
\end{equation}
where the input of the $l$$^{th}$ layer network is $H$$^{(l)}$ (the initial input is $H$$^{(0)}$), $N$ is the number of nodes in the graph, and each node is represented by the feature vector of the $D$ dimension. $\tilde{A}$=$A$+$I$$_N$ is added self-joining adjacency matrix, $\tilde{D}$ is a degree matrix,$\tilde{D}$$_{(ii)}$= $\Sigma_j$$\tilde{A}_{(ij)}$ . $W$$^{(l)}$$\in$$\mathbb{R}$$^{(D\times D)}$ is the parameter to be trained. $\sigma$ is the corresponding activation function.


\subsection{Transferring Illumination}

Our transfer network is based on a GAN. Its generator is an Multilayer Perceptron(MLP) consisting of 5 layers of FC. Except for the last layer, each layer is followed by a Batch Normalization(BN) layer and a LeakyReLU layer. The parameter of the LeakyReLU layer is 0.2. Our discriminator structure is similar to the graph autoencoder, consisting of two layers of convolution and two layers of FC layers. All layers except for the last layer are connected to the BN layer, and then all the convolution layers are connected to the tanh layer.

The generator transfers the latent feature vector of domain $B$ from that of the input domain $A$. The decoder obtain the OI of the virtual object. The discriminator determines whether the generated OI conforms to the distribution of domain $B$. Through this minimax game, the final generator produces properties of the real target object. In order to alleviate the mode collapse, we use the technique of Unrolled GAN~\cite{metz2016unrolled}. $G$ updates itself by predicting $D$'s future response in advance, making $D$ more difficult to respond to $G$'s update, and avoiding the problem of mode skipping.

\begin{figure*}
	\centering  
	\subfigure[]{
		\includegraphics[width=0.42\textwidth]{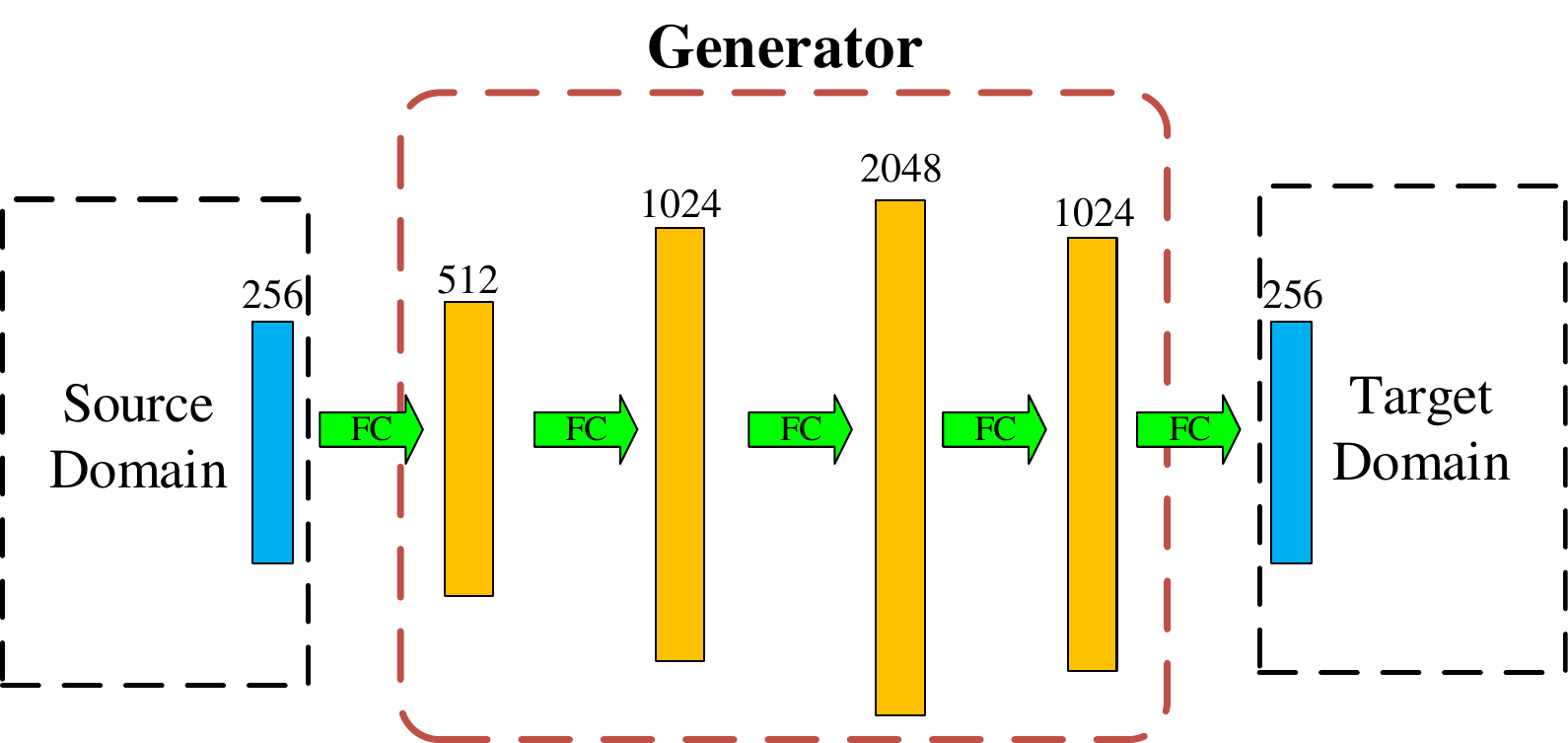}}
	\subfigure[]{
		\includegraphics[width=0.5\textwidth]{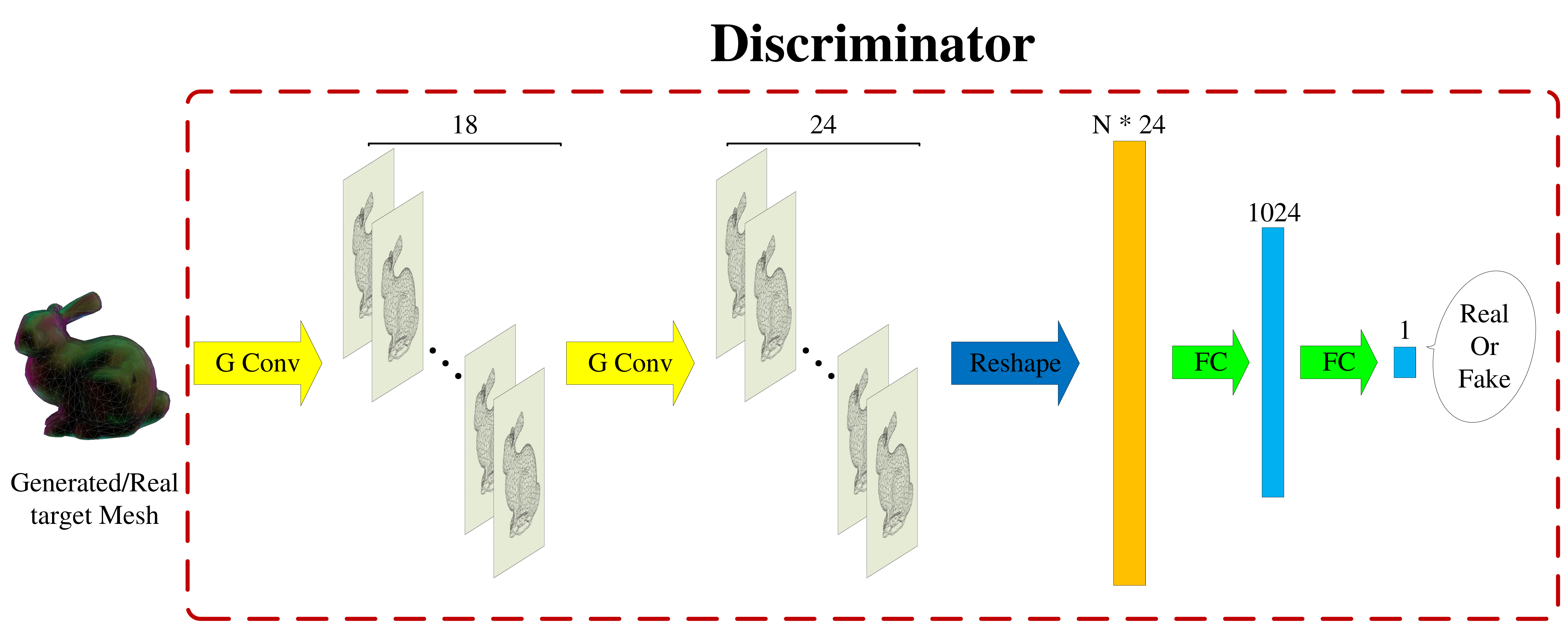}}
	\caption{(a)Generator: it translates the latent space vector of the source domain to the latent space of the target domain. (b)Discriminator: it discriminates whether the input data matches the distribution of the target domain.}
	\label{fig:gan}
\end{figure*}

\subsection{Color Rendering}
Since the nature of~\cite{xu2018shading} is 3D reconstruction, one can only infer OI from the color of 3D models, but not vice versa. Therefore we design another GAE that learns the color of 3D models from the corresponding OI. Its structure is almost the same as the GAE structure described in Sec.~\ref{sec:GAE}. The only difference is that this GAE generates the feature of $N$$\times$1, and then we will expand it to $N$$\times$3 to get the intensity value of the final object. Also, in order to cast the shadow correctly, the dominant OI region on the virtual object are first computed. Then the corresponding lighting direction is synthesized  based on the dominant OI region. Given the geometry of the virtual object and the plane underneath, we can finally cast the shadow accordingly.


\subsection{Loss Functions}

\hspace{0.2in}	\textbf{GAE}: Our GAE will eventually output the OI feature of N$\times$3. For the input 3D object $\mathcal{P}$, the network will reconstruct $\mathcal{\hat{P}}$. We define the reconstruction loss function of GAE shown in Eq.(2).:

\begin{equation}
L_{recons} = \frac{1}{ND} \sum_{i=0}^{N}\sum_{k=0}^{D} {\Big\lvert p_i^{k} - \hat{p}_i^{k} \Big\rvert }
\end{equation}
where $N$ is the number of points in the model and $D$ is the number of OI features, which is 3. $K$ represents the k$^{th}$ OI feature.

\textbf{GAN}: The initial objective function of our GAN is the squared error, that is, the essence of our transfer network is Least Squares Generative Adversarial Networks(LSGAN)\cite{lsgan17}. We define data($\textit{y}$) to represent the data of the target domain $\emph{T}$, i.e. $\textit{y}$ $\in$ $\emph{S}$, data($\textit{x}$) to represent the data of the source domain $\emph{S}$, i.e. $\textit{x}$ $\in$ $\emph{S}$. The real data is defined $\textit{y}$ as 1. The fake data is defined G($\textit{x}$) as 0. The the loss of GAN is defined in Eq.(3).:

\begin{equation}
\begin{aligned}
L_{LSGAN} = E_{y\thicksim{data}{(y)}} [(D(y)-1)^2] \\+ E_{x\thicksim{data}{(x)}} [(1-D(G(x)))^2]
\end{aligned}
\end{equation}

In order to generate the features of the corresponding target object from the source object, we also add the pairing loss represented in Eq.(4).:
\begin{equation}
L_{pair} = E_{ x\thicksim{data}{(x)} , y\thicksim{data}{(y)}} = [ \lvert y -G(x) \rvert]
\end{equation}

Moreover, we make use of the photo consistency by adding a shading term in Eq.(5).:
\begin{equation}
E_{sh} =\sum_{i=1}^{N}\|L(v_{i})\cdot {\bf n}(v_{i})-c_{i}\|^{2}
\end{equation}
where $E$ is the set of all edges of the mesh and $c_i$ is the
average of the intensity values in all
the multi-view images corresponding to vertex $v_i$. This is the intensity
error measuring the difference between the computed reflected
radiance and the average of the captured intensities.

Since the OI is supposed to be piece-wise smooth, we calculate the smooth loss of the 3D model, which is defined in Eq.(6).:
\begin{equation}
\bigtriangledown_M = \sum_{i=0}^{N} \sum_{k=0}^{D}{\Big\lvert \big(\frac{1}{d_i}\sum_{j\in N_i}{ p_j^k } \big)-p_i^k\Big\rvert}
\end{equation}
where \emph{d}$_i$ represents the degree of the i$^{th}$ node and \emph{N}$_i$ represents all neighbor nodes of the i$^{th}$ node.

Its matrix form is represented in Eq.(7).:
\begin{equation}
\bigtriangledown_M = {average}( D^{-1}AM-M )
\end{equation}
where \emph{D} is the degree matrix, \emph{A} is the feature matrix, and \emph{M} is the adjacency matrix.

Finally, the total loss can be calculated in Eq.(8).:
\begin{equation}
\begin{aligned}
L_{total} = L_{LSGAN}+\beta_{pair}L_{pair}\\+\beta_{shading}E_{sh}+\beta_{smooth}\bigtriangledown_M
\end{aligned}
\end{equation}

%
%
%

%

\subsection{Implementation Details}
\hspace{0.2in}	    \textbf{Dataset}: We first generate the synthetic data for training. A total of $10,000$ sets of synthetic lighting environment are generated randomly to model the indoor and outdoor illuminations. For each lighting condition, 32 synthetic point light sources are randomly placed in the 3D space. The corresponding OI and intensity of the virtual object are then computed. Additionally, a rotation perturbation is applied to the object, so that our GAE can learn feature representation with various poses.


For real-world dataset, we use Debevec's mediancut algorithm~\cite{debevec2008rendering} to generate $3,292$  real environment illuminations from the real-world HDR environment maps of SHlight\cite{Cheng2018}, Laval Indoor Dataset\cite{gardner-sigasia-17}, which are represented as 32-point sources. We cropped the LDR image with random pitch, yaw and exposure, and got the HDR lighting under this setting. We use GT lighting and predicted lighting of previous methods to render the target object respectively. There are number of $247$ sets of data used as the test split. Then we randomly rotated these point sources three times to get $9,135$ augmented data, among which $1,000$ (9$\%$) are used as validation data. Our algorithm generates the corresponding OI and intensity for different objects as our fine-tuning data.

\textbf{Training}: The training is conducted with four GTX 1080TI GPUs, and the whole procedure takes around four hours. We train the GAE of the planar surfaces and that of the virtual object separately. After that the renderer of the virtual object is trained, and finally the transfer network. GAE and the renderer are trained them for $400$ epochs with a batch size of $256$, using the ADAM optimizer with betas of $(0.9, 0.999)$ and learning rate of $0.001$.  For the transfer network, we set the ADAM optimizer with betas of $(0.5, 0.99)$, $G$ and $D$ with the learning rate of $0.0001$ and 0.0004, respectively, and train for $100$ epochs. $\beta_{pair}$ is 1.0, $\beta_{smooth}$ is 2.5, and $\beta_{shading}$ is 0.3. For all dropout layers in the GCN layer, the parameter is 0.2. 

\begin{table*}
	\centering \caption{Quantitative comparison between the state-of-the-arts and our method on relighting errors using real-world data. Note that the relighting error is computed in 3D space, where all the vertices on the virtual object are considered.}\label{tb:evaluate} 
	\begin{tabular}{|l|c|c|c|c|c|c|} \hline
		&\multicolumn{2}{|c|}{Indoor}
		&\multicolumn{2}{|c|}{Outdoor}
		&\multicolumn{2}{|c|}{Spatially-varying}  \\ \cline{2-7}
		&MAE   &RMSE   &MAE   &RMSE  &MAE   &RMSE     \\ \hline
		
		~\cite{Chloe19} &0.122   &0.151   &0.116 &0.143 &N.A.&N.A. \\\hline
		~\cite{gardner-sigasia-17} &0.142 & 0.179    &0.145 &0.175 &N.A.&N.A.  \\\hline
		~\cite{hold2019deep}  &N.A.   &N.A.   &0.109 &0.132 &N.A. &N.A. \\\hline
		~\cite{hold2017deep}  &N.A.   &N.A.   &0.159 &0.202 &N.A.&N.A. \\\hline
		~\cite{Garon_2019_CVPR}  &N.A.   &N.A.   &N.A. &N.A. &0.072 &0.089 \\\hline
		Ours  & \textbf{0.066} &\textbf{0.081}  &\textbf{0.061} &\textbf{0.076 }&\textbf{0.056} &\textbf{0.070} \\\hline
	\end{tabular}
\end{table*}

\begin{table*}
	\centering \caption{Effects of different losses. The our proposed shading loss and smooth loss improve the relighting results. }\label{tb:ablation} 
	\begin{tabular}{|l|c|c|c|c|c|c|} \hline
		&\multicolumn{2}{|c|}{Indoor}
		&\multicolumn{2}{|c|}{Outdoor}
		&\multicolumn{2}{|c|}{Spatially-varying}  \\ \cline{1-7}
		Loss terms  &MAE   &RMSE   &MAE   &RMSE  &MAE   &RMSE     \\ \hline
		
		Original &0.066 & 0.082    &0.065 &0.081 &0.059 &0.073  \\\hline

		$L_{shading} + L_{smooth}$ &0.066   &0.081   &0.061 &0.076 &0.056 &0.070 \\\hline
	\end{tabular}
\end{table*}

\begin{figure*}
	\begin{center}
		\includegraphics[width=0.95\textwidth]{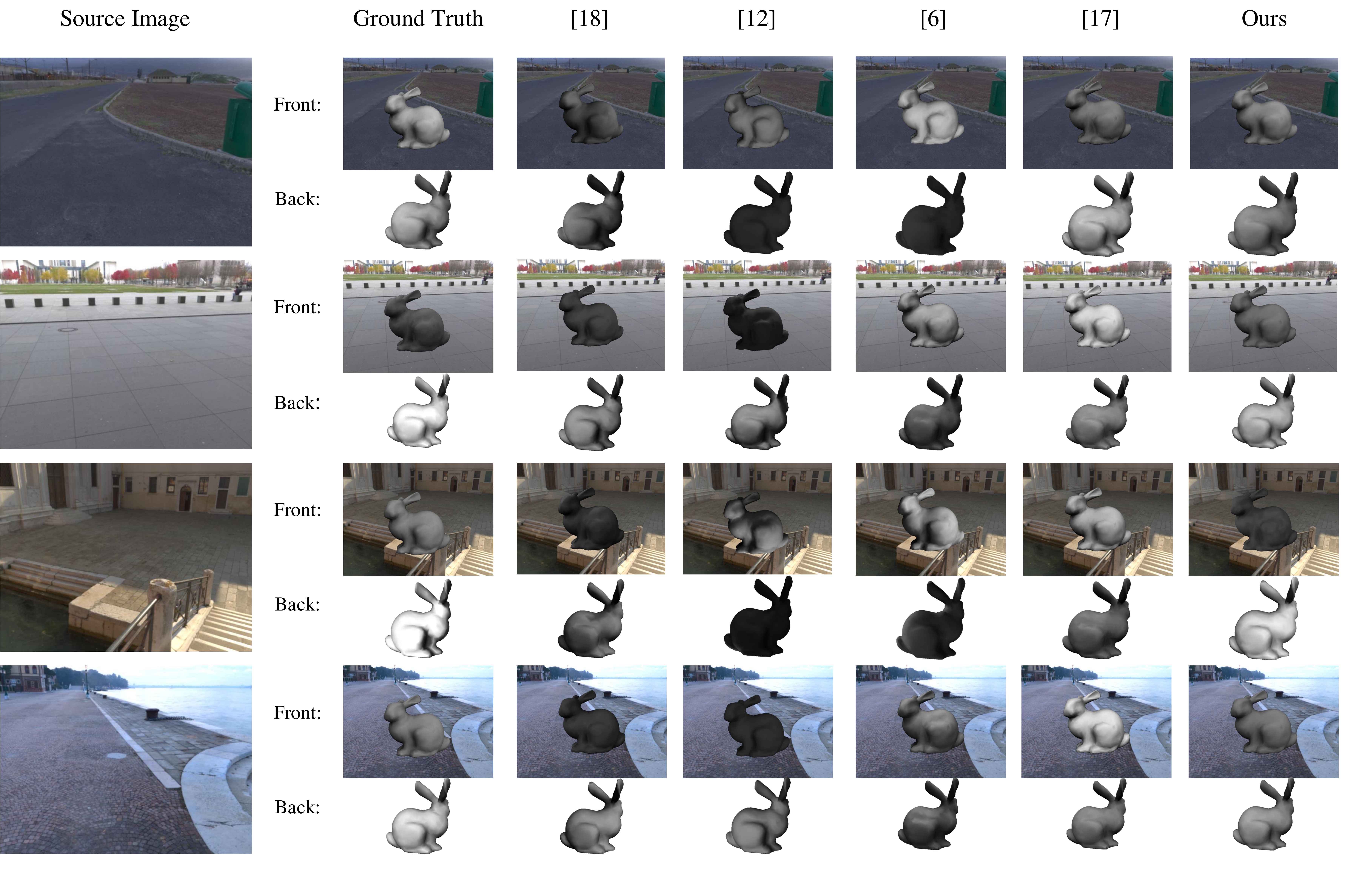}
	\end{center}
	\caption{Outdoor results. From left to right: ground truth, results from~\cite{hold2017deep}, ~\cite{gardner-sigasia-17}, ~\cite{Chloe19}, ~\cite{hold2019deep}
		and our results. Note that some models may look realistic from the frontal-view, but their back-views are quite different from the ground truth.}
	\label{fig:outdoor}
\end{figure*}

\begin{figure*}
	\begin{center}
		\includegraphics[width=0.8\textwidth]{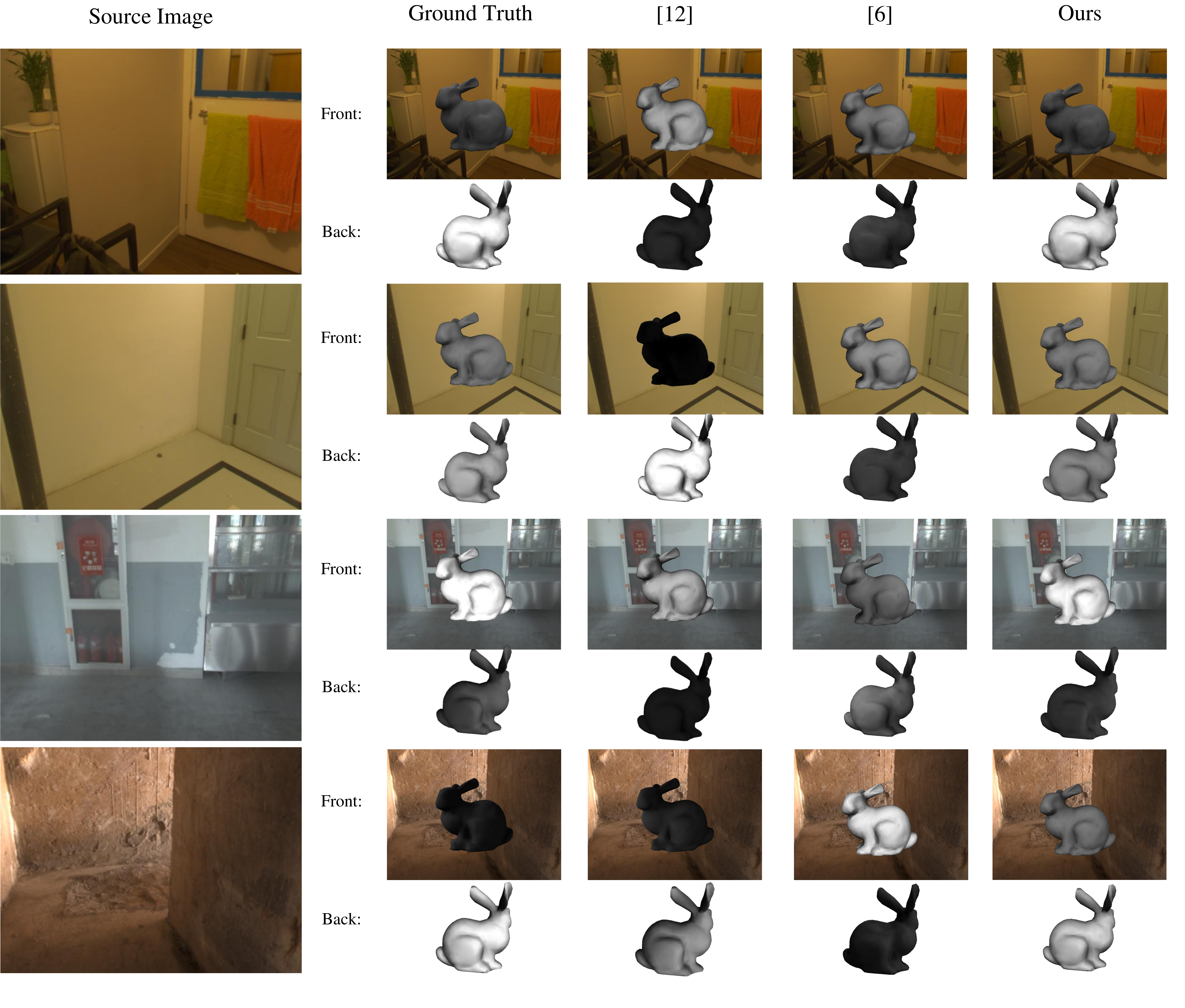}
	\end{center}
	\caption{Indoor results. From left to right: ground truth, results from~\cite{gardner-sigasia-17}, ~\cite{Chloe19}, and our results. Note that some models may look realistic from the frontal-view, but their back-views are quite different from the ground truth. }
	\label{fig:indoor}
\end{figure*}

\begin{figure*}
	\begin{center}
		\includegraphics[width=0.9\textwidth]{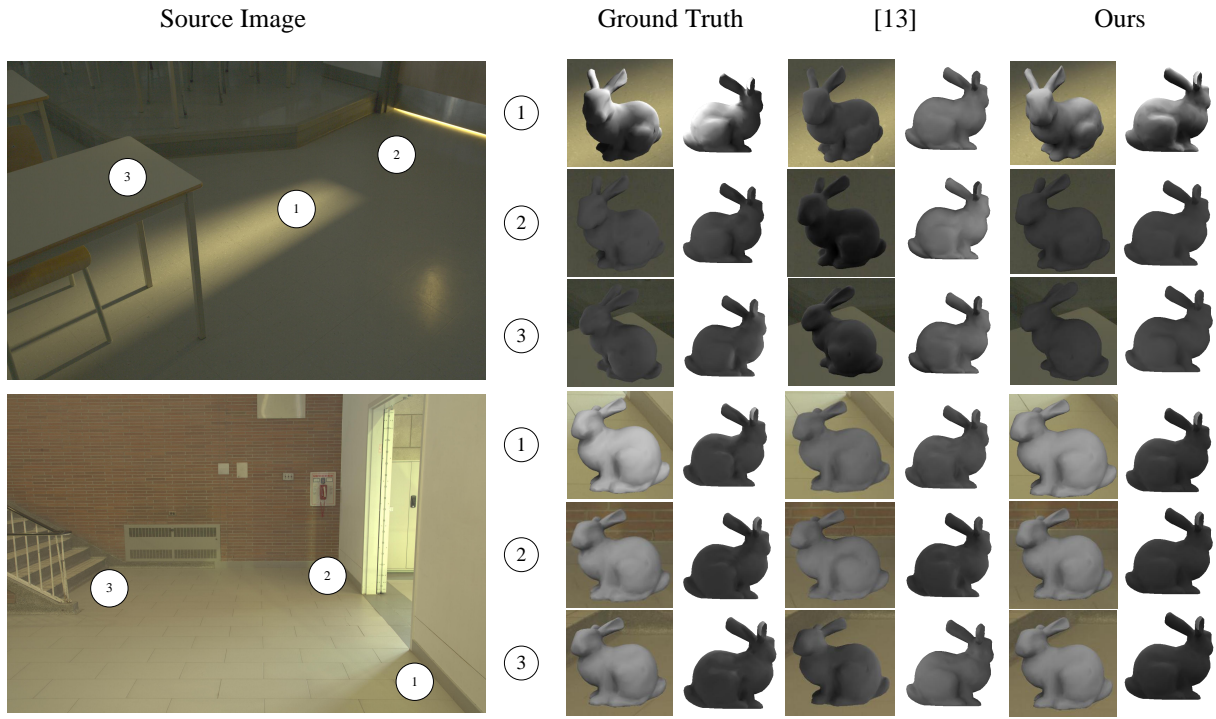}
	\end{center}
	\caption{Spatially-varying results. The numbers indicate different rendering positions of the virtual object. From left to right: source image with position marks, ground truth, results from~\cite{Garon_2019_CVPR} and our results.}
	\label{fig:varying}
\end{figure*}

\section{Experiment}
Our proposed method is evaluated quantitatively and qualitatively on several test sets. To show the robustness of our algorithm, extensive comparisons are conducted to state-of-the-arts that proposed for different scenarios, namely, indoor~\cite{gardner-sigasia-17,Chloe19}, outdoor~\cite{hold2017deep,hold2019deep,Chloe19} and spatially-varying data~\cite{Garon_2019_CVPR}. There are totally $1,000$ sets of testing lighting conditions in our synthetic data. For the real-world data, there are 141 indoor lighting scenes from SHlight~\cite{Cheng2018} and Laval Indoor Dataset~\cite{gardner-sigasia-17}, 106 outdoor lighting scenes from SHlight~\cite{Cheng2018}, and 76 lighting scenes from spatially-varying data in~\cite{Garon_2019_CVPR}. In cases that planes are failed to be detected, we place a synthetic planes in the image as the source for our feature transferring framework.  


\subsection{Quantitative Results}
We evaluate the aforementioned reported in literature and ours by computing the relighting errors of the virtual object. It's worth mentioning that previous works only account for the relighting error from a single view. That is to say, these prior works crop a 2D image from a particular view with the virtual object in the scene, and calculate its relighting error on a pixel-wise basis. We argue that measuring the rendered objects in 3D space is more reasonable. This is because that nowadays the multi-user AR applications are becoming more and more common. For example, Microsoft's \emph{Azure Spatial Anchors}~\cite{anchors} enables multiple users to place virtual content in the same physical location, where the rendered objects can be seen on different devices in the same position and orientation relative to the environment. In such a case, pixel-wise measurement from a single view is not enough. Given the nature of our object relighting framework, we can directly measure the 3D relighting error of our result.

%

As shown in Table~\ref{tb:evaluate}, our method outperforms state-of-the-arts on all the three types of data. In particular, our method improves substantially compared to the indoor~\cite{gardner-sigasia-17,Chloe19} and outdoor~\cite{hold2017deep,hold2019deep,Chloe19} methods. 
As for the  spatially-varying data,  our error is also lower than~\cite{Garon_2019_CVPR}. The overall performance on all datasets also shows the robustness of our method. Table~\ref{tb:ablation} shows the shading loss and smooth loss of our proposed approach improve the relighting results. This is because that the shading loss enforces the photo-consistency of the rendered virtual object, based on its geometry and lighting. And the smooth loss means that the lighting is expected to be piece-wise smooth.

%


\subsection{Qualitative Results}

We provide qualitative results of all methods in these scenarios, which are shown in the Fig.~\ref{fig:outdoor}, ~\ref{fig:indoor}, and~\ref{fig:varying}.  
As mentioned earlier, our framework relights the object from all directions in 3D space, instead of a single view. Therefore, we also show the back-view of the rendered model in all scenes. We'd like to point out that this step is proven to be quite important. As some methods may perform well on the front, their back-view is not realistic. This problem is particular obvious for Deeplight~\cite{Chloe19}, which can be seen in both Fig.~\ref{fig:outdoor} and~\ref{fig:indoor}, as their back-view results look relatively dark compared to the Ground Truth(GT) or other methods. We believe that this may be caused by their training data, which were captured with the mobile phone camera from a very close distance to the light probe. In such a set-up, if there is a strong light source, e.g. the sun, locating just behind the light probe seeing from the camera view, the generated environment map would fail to record this light source. This may explain why the back-views of their results are relatively dark. It shows that capturing HDR environment maps with light probes may bring problems which have been overlooked by previous works.

Meanwhile, observed from Fig~\ref{fig:outdoor}, outdoor methods based on sun and sky-model~\cite{hold2017deep,hold2019deep} are difficult to generate satisfactory results if the sun is not seen from the input image. This is due to the nature of their methods. As a matter of fact, there are many outdoor images captured without the sun or sky. So this is one limitation of such methods. Especially in the third row, as discussed in Fig.7 \& 8 of ~\cite{Cheng2018}, we observed that the back-view of GT is bright, because there is a strong light source, sun, above the building. .

Similar with~\cite{Garon_2019_CVPR}, our method also has the spatially-varying capability. For different locations on a same image, the rendered model appears differently according to its relative position to the light sources. Although our improvement may not look so significant compareed to the results of ~\cite{Garon_2019_CVPR} in Fig.~\ref{fig:varying}, their method is meant for indoor scenes only. This makes our method more robust as it works in outdoors as well.

We further conduct a user study to evaluate the realism of results. Users were shown pairs of images with inserted objects and asked to pick the more realistic ones. Each pair was either rendered with GT lighting or the prediction from one of~\cite{Chloe19,Garon_2019_CVPR,hold2019deep}. A total number of 170 unique participants took part in the study, and 17 scenes with inserted virtual objects were given. Results are given as percentages, denoting the fraction that each method was preferred to the GT illumination (the higher the better). For spatially-varying data (e.g. Fig.~\ref{fig:varying}), our method achieved 32.5\% , compared to 28\% for~\cite{Garon_2019_CVPR}. For the rest data, our method achieved 48.8\%, compared to 39.6\% for~\cite{Chloe19} and 41.4\% for~\cite{hold2019deep}. Overall, users had a higher preference for our predictions.

%



\begin{figure*}
	\begin{center}
		\includegraphics[width=0.9\textwidth]{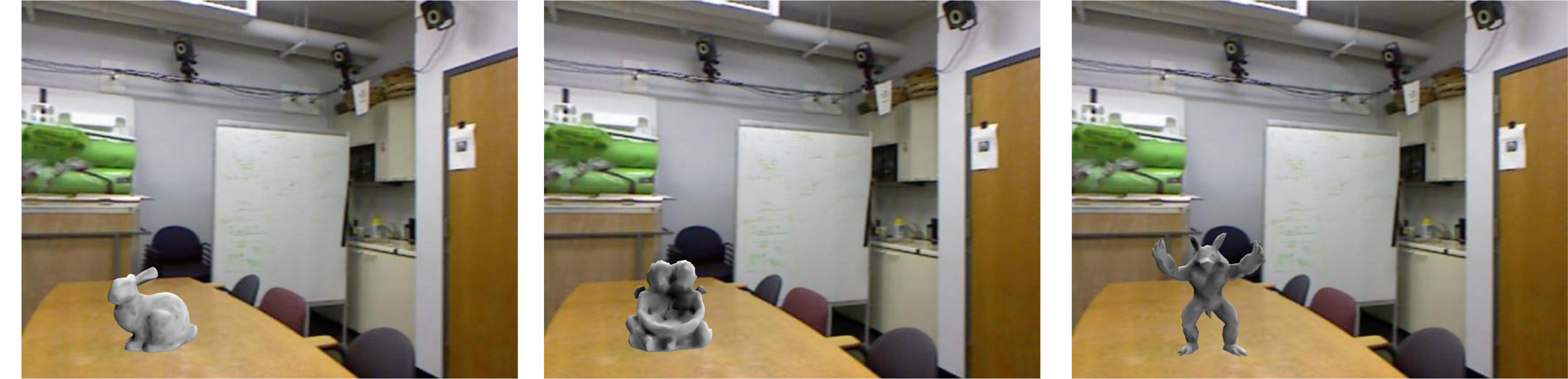}
	\end{center}
	\caption{Different virtual models rendered in the same scene. Since our framework is geometry-based, each virtual model needs to be retrained. }
	\label{fig:diff3d}
\end{figure*}

\vspace{0.1in}
\textbf{Limitations}: Since our framework is geometry-based, it requires retraining for each new type of virtual objects. For each 3D model shown in Fig.~\ref{fig:diff3d}, the training takes around four hours on our server. However we'd like to mention that for most AR applications, the virtual objects are already installed or pre-defined. That is to say, the geometries are known in advance and offline training is practical. 

%

\section{Conclusion}
We present a novel algorithm for virtual object illumination estimation. Instead of reconstructing the lighting of the entire real scene, we directly transfer the illumination effects from existing planar surfaces to the virtual object. Our feature transferring algorithm is based a GAN, with plane detection and OI estimation as pre-processing steps. Extensive experiments have been conducted on indoor, outdoor, and spatially-varying data. It is shown that our method can accurately estimate the illumination of virtual objects in real scenes.

\ifCLASSOPTIONcompsoc
%
%

\ifCLASSOPTIONcaptionsoff
  \newpage
\fi



%
%
%
\bibliographystyle{IEEEtranS}
\bibliography{egbib}

\end{document}